\documentclass[preprint,5p,times,twocolumn]{elsarticle}

\usepackage{amssymb}  % provides various useful mathematical symbols
\usepackage{amsmath}  % provides mathematical formulas
\usepackage{hyperref}  % provides hyper links to references, figures, and tables
\usepackage{caption}
\usepackage{upgreek}  % For getting upright greek letters, using $\uptheta$
\usepackage{multirow}  % For using \multirowcell in table
\usepackage{makecell}  % For using \multirowcell in table

\journal{Journal of Electronic Imaging}

\begin{document}

\begin{frontmatter}

\title{Label distribution based facial attractiveness computation by deep residual learning}

\author[mymainaddress]{Shu Liu\corref{euqalauthor}}
\cortext[euqalauthor]{Parallel first author}

\author[mymainaddress]{Bo Li\corref{euqalauthor}}
\ead{libo.npu@gmail.com}

\author[mymainaddress]{Yangyu Fan}

\author[mymainaddress]{Zhe Guo}

\author[mysecondaryaddress]{Ashok Samal}

\address[mymainaddress]{School of Electronics and Information, Northwestern Polytechnical University, Xi'an, Shaanxi, China}
\address[mysecondaryaddress]{Department of Computer Science and Engineering, University of Nebraska-Lincoln, Lincoln, NE, USA}

%%%%%%%%%%%%%%%%%%%%%%%%%%%%%%%%%%%%%%%%%%%%%%%%%%%%%%%%%%%%%
\begin{abstract}
Two challenges lie in the facial attractiveness computation research: the lack of true attractiveness labels (scores), and the lack of an accurate face representation. In order to address the first challenge, this paper recasts facial attractiveness computation as a label distribution learning (LDL) problem rather than a traditional single-label supervised learning task. In this way, the negative influence of the label incomplete problem can be reduced. Inspired by the recent promising work in face recognition using deep neural networks to learn effective features, the second challenge is expected to be solved from a deep learning point of view. A very deep residual network is utilized to enable automatic learning of hierarchical aesthetics representation. Integrating these two ideas, an end-to-end deep learning framework is established. Our approach achieves the best results on a standard benchmark SCUT-FBP dataset compared with other state-of-the-art work.
\end{abstract}


\end{frontmatter}

%%%%%%%%%%%%%%%%%%%%%%%%%%%%%%%%%%%%%%%%%%%%%%%%%%%%%%%%%%%%%
\section{Introduction}
The attractiveness of a face retains influence on many social endeavors. Beautiful faces can have effects on person's personality, career prospect, and their personal relationships \cite{OHFP11_Calder}. Integrating machine learning and computer vision techniques, automatic facial attractiveness computation has become an ever-growing topic.

\begin{figure}[!h]
\centering
\includegraphics[width=90mm]{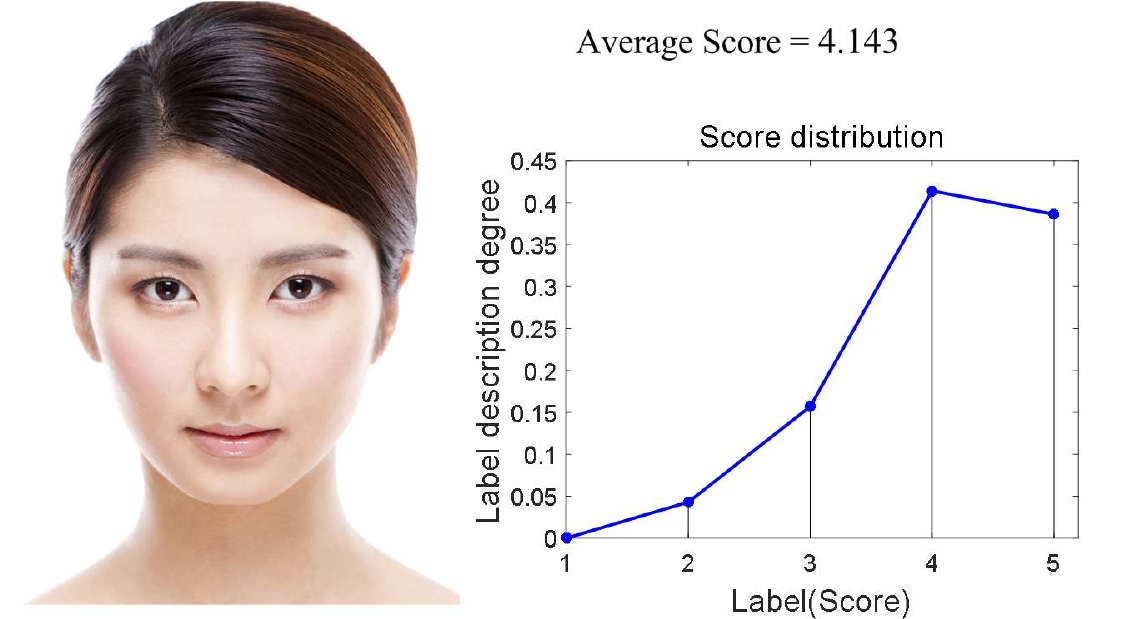}
\caption{The average attractiveness score and score distribution of one face in the SCUT-FBP dataset \cite{SMC15_Xie}.}
\label{fig1}
\end{figure}

Although substantial progress has been achieved in the research of face attractiveness computation, challenges still remain. The first challenge is the lack of true attractiveness labels (scores). In order to derive an approximation of the ground truth, a typical way is to survey a diverse group of human raters who assign scores to a set of faces. The average score of each face is then defined as the ground-truth label for the following classification or regression task \cite{myMTAP}. However, the average score is not always a good indicator of universally accepted preference, especially for the controversial faces. On the contrary, the score distribution collected from different raters provides more aesthetics-degree information of a face than a single label. Fig. \ref{fig1} gives an example of the average attractiveness score versus score distribution of one face. We can easily find that the score distribution includes the distribution of a number of labels, and represents the extent to which each beauty level describes the overall attractiveness of the face. In this sense, the score distribution can be viewed as a natural representation of a label distribution. Therefore, we recast facial attractiveness computation as a Label Distribution Learning (LDL) problem \cite{PAMI13_Geng}, a newly proposed machine learning paradigm. The LDL is able to deal with insufficient and incomplete training data, since each face is expected to contribute to the learning of a number of attractiveness levels.

The second challenge is the lack of an accurate face representation that captures the salient elements of attractiveness. Many heuristics rules have been quantitatively studied over the years, inspiring researchers to hand-design diverse features in the most previous studies. The features can be geometric, color, texture based \cite{NC06_Eisenthal,NIPS06_Kagian,PR08_Schmid}, as well as appearance descriptors \cite{ECCV10_Gray,PRIACVA12_Bottino} either at local or holistic scale. Recently, the up-to-date deep learning methods, especially convolutional neural network (CNN), have been applied to automatically learn a hierarchical and higher-level face representation for facial attractiveness computation task \cite{Neucom14_Gan,arXiv15_Xu}. Even with the shallow and plain architecture, i.e., two-layer convolutional restricted Boltzmann machine in \cite{Neucom14_Gan} and six-layer CNN in \cite{arXiv15_Xu}, superior performance has been achieved to the previous work with traditional hand-crafted features. Since face beauty is a complex concept with no universal-accepted representation, we are motivated to determine the most appropriate visual characteristics from the raw RGB image by very deep networks.

In this work, we intend to explore the deeper facial aesthetics features by learning features from raw images directly through deeper architecture. A very deep convolutional residual network (ResNet) \cite{CVPR16_ResNet} is utilized which takes RGB pixels as inputs and automatically learn an effective face representation. We also incorporate the idea of LDL and ResNet in facial attractiveness computation, and show the advantages of our method on a newly-constructed benchmark dataset SCUT-FBP \cite{SMC15_Xie}.

%%%%%%%%%%%%%%%%%%%%%%%%%%%%%%%%%%%%%%%%%%%%%%%%%%%%%%%%%%%%%
\section{Our method}
\subsection{Network architecture}
In order to use deep convolutional residual network \cite{CVPR16_ResNet}, the input images are need to normalized to a fixed size, i.e., $224 \times 224 \times 3$ in this work. We first extract face region from the original image, making it possible to concentrate solely on the attractiveness of the face itself. We then pad the border pixels with zeros to the shorter side of the image to generate a normalized square input of our network.

\begin{figure}[!h]
\centering
\includegraphics[width=85mm]{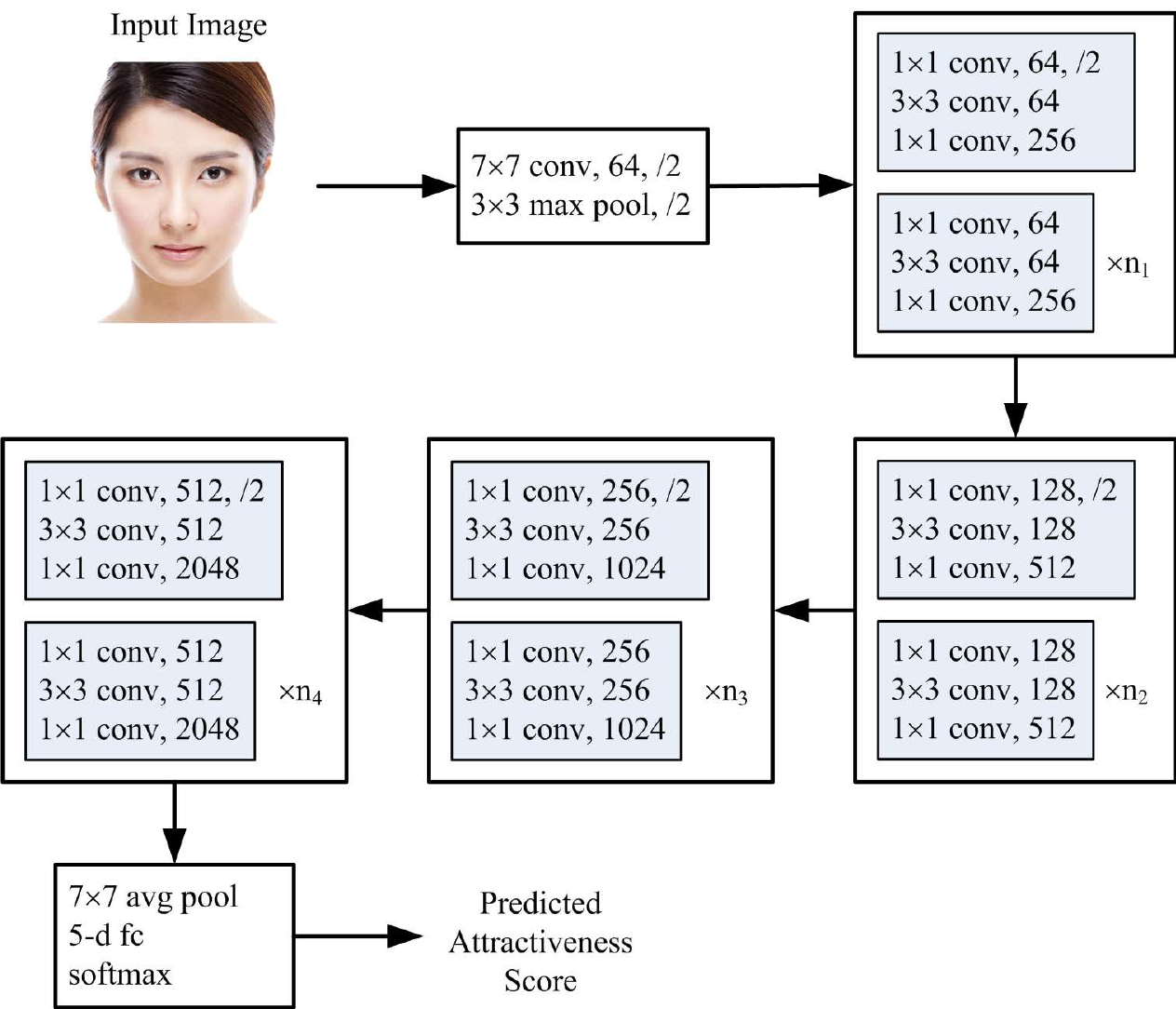}
\caption{The network architecture of our model.}
\label{fig2}
\end{figure}

The overall architecture of the ResNet used for facial attractiveness computation is shown in Fig. \ref{fig2}. It contains a convolutional layer, a max-pooling layer, and four convolutional ``bottleneck" blocks followed by an average-pooling layer, a fully-connected layer with 5 neurons, and a softmax layer. Batch normalization is performed after each convolution layer. With different value of $n_1$, $n_2$, $n_3$, $n_4$ in the bottleneck blocks, the residual network architectures can be constructed into different depth. More details about the architecture can be found in \cite{CVPR16_ResNet} In this work, we consider the 50-layer and 101-layer ResNet where \{$n_1$,$n_2$,$n_3$,$n_4$\} is \{2,3,5,2\} and \{2,3,22,2\}, respectively.

%%-----------------------------------------------------------
\subsection{Loss function}
While the existing work uses a single-label (average score) for supervised regression, we consider the score distribution to describe the attractiveness of a face, which can be viewed as a natural representation of a label distribution. In this work, therefore, we recast facial attractiveness computation as a LDL problem.

Similar to the definitions in \cite{PAMI13_Geng}, suppose for the $i$-th image, the feature representation extracted from 5d-fc layer is $\mathbf{x}_i$, the complete set of possible labels (scores) is $\mathbf{y}=\{y_1,y_2,\ldots,y_c\}$, and the score distribution is $\mathbf{d}_i=\{d^{y_1}_{\mathbf{x}_i},d^{y_2}_{\mathbf{x}_i},\ldots,d^{y_c}_{\mathbf{x}_i}\}$ where $d^{y_j}_{\mathbf{x}_i}$ denotes the description degree (distribution) of the label $y_j$ to the instance $\mathbf{x}_i$ and $\sum_{j}{d^{y_j}_{\mathbf{x}_i}}=1$. Given a training set $\{(\mathbf{x}_i,\mathbf{d}_i),1\leq i \leq n\}$, the purpose is to train a set of parameters $\boldsymbol{\uptheta}$ to generate a distribution $f(\mathbf{x}_i;\boldsymbol{\uptheta})$ similar to $\mathbf{d}_i$. Here the Euclidean distance and Kullback-Leibler divergence are used as the measurement of the similarity of these two distributions. The training of the last softmax layer is done by minimizing the either of the following overall loss functions:

\begin{equation}
\label{eq1}
L_{Eu}=\sum_{i}{\lVert{\mathbf{d}_i-f(\mathbf{x}_i;\boldsymbol{\uptheta})}\lVert_2}
\end{equation}

\begin{equation}
\label{eq2}
L_{KL}=\sum_{i}{\mathbf{d}_i\ln{\frac{\mathbf{d}_i}{f(\mathbf{x}_i;\boldsymbol{\uptheta})}}}
\end{equation}

Given a new test image, the normalized input and its patch is fed into the network to compute the feature representation $\mathbf{x'}$, and the predicted label distribution can be generated by $f(\mathbf{x'};\boldsymbol{\uptheta})$, which provides the distribution of attractiveness scores. We can also predict the exact score of this face by weighted mean of the score distribution.

%%%%%%%%%%%%%%%%%%%%%%%%%%%%%%%%%%%%%%%%%%%%%%%%%%%%%%%%%%%%%
\section{Experiments}
We evaluate our proposed method on the SCUT-FBP dataset \cite{SMC15_Xie}, which contains 500 facial images with around 70 attractiveness scores from 1-5. Following the data partition setting in \cite{arXiv15_Xu}, 400 images are randomly selected as the training set, and the remaining 100 images as the testing set. Since the comparable work aims at exact attractiveness estimation, we calculate the weighted mean score $\hat{y'}$ from the predict distribution by $\hat{y'}=f(\mathbf{x'};\boldsymbol{\uptheta})\cdot\mathbf{y}$. A Pearson correlation (PC) between $\hat{y'}$ and ground-truth average score from raters is used to evaluate the performance of our model.

Our experiments are conducted on the deep learning platform of Caffe \cite{ACMM14_Caffe}. We fix the batch size as 32, and weight decay as 0.0005. The learning rate is initiated with 0.001, and reduced by a factor of 10 for every 4,000 iterations. The max iteration is set to 17,000. The learning rate of the last layer is increased by a factor of 10 for speeding up the convergence, and weight decay is multiplied by 100 to avoid overfitting.

\begin{figure}[!h]
\centering
\includegraphics[width=90mm]{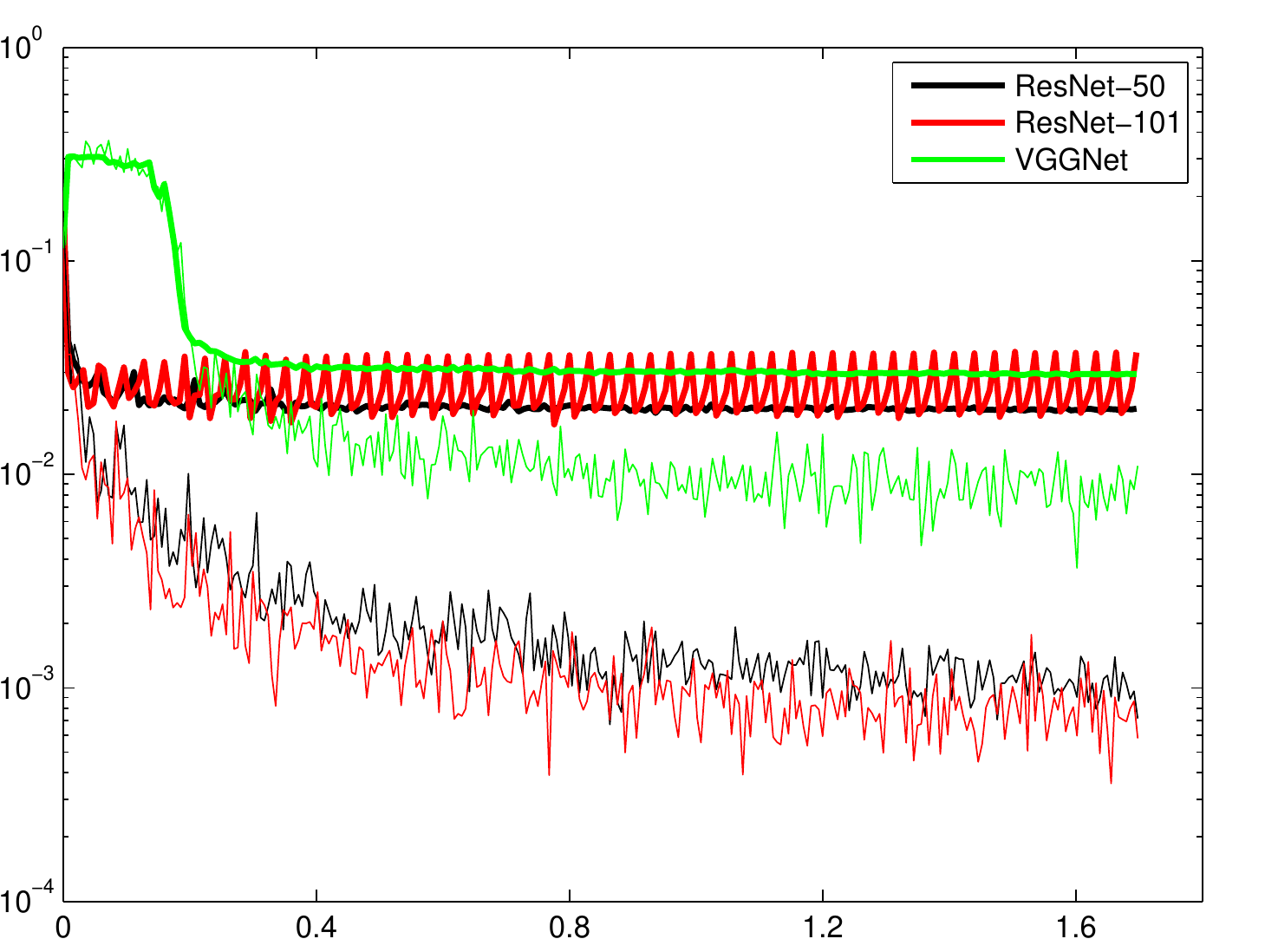}
\caption{Training behavior on three networks. Thin curves denote training error, and bold curves denote test error.}
\label{fig3}
\end{figure}

To start with, we identify the role of deeper networks in facial attractiveness computation in the traditional regression context. Due to the very limited number of training samples, we fine-tune on three classic models including ResNet-50, ResNet-101 and VGGNet-19 \cite{ICLR15_VGGNet}, which were pre-trained on the Imagenet dataset \cite{CVPR09_Imagenet}.
The performance of three networks is compared in Table \ref{tab1}. We need to mention that the baseline result in \cite{arXiv15_Xu} is the correlation of 0.83 rather than the highest 0.88 achieved by several inputting channels instead of raw images. All the three networks achieve superior performance to the baseline because of the much deeper architectures. The 50-layer ResNet performs best, and VGGNet worst. To reveal the possible reasons, Fig. \ref{fig3} presents the behaviors of three networks throughout the training procedure. Without the residual structure, VGGNet cannot be effectively trained due to the degradation problem which was described in \cite{CVPR16_ResNet}, thus the training error of VGGNet is the highest during the whole training. Compared to ResNet-50, the inferior performance of 101-layer ResNet is caused by overfitting, leading to the lower training error and wavy testing error as shown in Fig. \ref{fig3}. It is notable that we experiment Euclidean distance and KL divergence in the loss function, where Euclidean distance achieves slightly better. Therefore, all the results are obtained under this setting.

\begin{table}[!hbp]\footnotesize
%\centering
\caption{Comparison of different networks.}
\label{tab1}
\begin{tabular}{|l|l|l|l|l|}
\hline
Networks & \cite{arXiv15_Xu} & ResNet-50 & ResNet-101 & VGG-19 \\
\hline
PC & 0.83 & 0.87 & 0.85 & 0.84 \\
\hline
\end{tabular}
\end{table}

With the network fixed to ResNet-50, we then identify the role of label distribution in facial attractiveness computation. As shown in Table \ref{tab2}, by introducing LDL to our task, the correlation is increased by 2\%. It reinforces the advantage of label distribution that provides more aesthetics-degree information of than the average score. This framework potentially allows a more universally-accepted attractiveness score in contrast to both ethnically and view's gender-tuned model. In order to further boost attractiveness computation accuracy, several augmentation techniques, including the standard color augmentation \cite{NIPS12_AlexNet}, rotation, contrast enhancement, etc., are used to enlarge the size of the training data to 8,000. In this way, the highest correlation of 0.92 has been achieved.

\begin{table}[!hbp]\footnotesize
%\centering
\caption{Improvement of ResNet-50.}
\label{tab2}
\begin{tabular}{|l|l|l|l|}
\hline
\multirowcell{3}{Improvements} & \multirowcell{3}{ResNet-50} & \multirowcell{2}{ResNet-50} & ResNet-50 \\
& & & + LDL \\
& & + LDL & + Augmentation \\
\hline
PC & 0.87 & 0.89 & 0.92 \\
\hline
\end{tabular}
\end{table}

%%%%%%%%%%%%%%%%%%%%%%%%%%%%%%%%%%%%%%%%%%%%%%%%%%%%%%%%%%%%%
\section{Conclusion}
The purpose of this paper is to present a new deep learning based framework for facial attractiveness computation. Rather than using the average attractiveness score as the ground truth for single-label supervised learning, we recast this task as a label distribution learning (LDL) problem. In order to extract aesthetic-related face representation, a very deep residual network is utilized. Comprehensive experiments are performed on the SCUT-FBP dataset, where our approach achieves significant better results than the current state of the art. Based on this work, we believe that the construction of large-scale benchmark and more effective deep networks deserve attention in the future.

%%%%%%%%%%%%%%%%%%%%%%%%%%%%%%%%%%%%%%%%%%%%%%%%%%%%%%%%%%%%%
%\section*{References}
%\bibliography{mybibfile}

%  First use bib file, run the tex and generate bbl file. Then copy all the bibitem from it.

\end{document}